\documentclass{article}

    \usepackage[preprint]{neurips_2024}

\usepackage[utf8]{inputenc} % allow utf-8 input
\usepackage[T1]{fontenc}    % use 8-bit T1 fonts
\usepackage{hyperref}       % hyperlinks
\usepackage{url}            % simple URL typesetting
\usepackage{booktabs}       % professional-quality tables
\usepackage{amsfonts}       % blackboard math symbols
\usepackage{nicefrac}       % compact symbols for 1/2, etc.
\usepackage{microtype}      % microtypography
\usepackage{xcolor}         % colors
\usepackage{times}
\usepackage{latexsym}

\usepackage[T1]{fontenc}
\usepackage[utf8]{inputenc}

\usepackage{microtype}

\usepackage{inconsolata}
\usepackage{amsmath}
\usepackage{booktabs}
\usepackage{tikz}

\title{Formatting Instructions For NeurIPS 2024}

\begin{document}

\title{OpenStaxQA: A multilingual dataset based on open-source college textbooks}

% Author information can be set in various styles:
% For several authors from the same institution:
% \author{Author 1 \and ... \and Author n \\
%         Address line \\ ... \\ Address line}
% if the names do not fit well on one line use
%         Author 1 \\ {\bf Author 2} \\ ... \\ {\bf Author n} \\
% For authors from different institutions:
% \author{Author 1 \\ Address line \\  ... \\ Address line
%         \And  ... \And
%         Author n \\ Address line \\ ... \\ Address line}
% To start a separate ``row'' of authors use \AND, as in
% \author{Author 1 \\ Address line \\  ... \\ Address line
%         \AND
%         Author 2 \\ Address line \\ ... \\ Address line \And
%         Author 3 \\ Address line \\ ... \\ Address line}

\author{Pranav Gupta \\
\textit{Lowe's}}

\maketitle
\begin{abstract}
We present OpenStaxQA, an evaluation benchmark specific to college-level educational applications based on 43 open-source college textbooks in English, Spanish, and Polish, available under a permissive Creative Commons license. We finetune and evaluate large language models (LLMs) with approximately 7 billion parameters on this dataset using quantized low rank adapters (QLoRa). Additionally we also perform a zero-shot evaluation on the AI2 reasoning challenge dev dataset in order to check if OpenStaxQA can lead to an improved performance on other tasks. We also discuss broader impacts relevant to datasets such as OpenStaxQA.  
\end{abstract}

\section{Introduction}

Large language models (LLMs) used for STEM education suffer from a lack of high quality specialized training and evaluation datasets. This problem is aggravated for the case of languages other than English. Existing benchmarks such as AP tests and the AI2 reasoning challenge have focused on concepts typically at grade school level. Such high-quality datasets have been rarer at undergraduate or graduate course level, and training and evaluation datasets have mainly relied on online help formus such as Stack Exchange \cite{beeching2023stackllama}. Therefore, there is a need for creating high-quality datasets for training and evaluating models for STEM education applications. However, high-quality datasets in the form of online textbooks are typically unavailable for web scraping due to legal concerns.  While several educators regularly upload their textbooks online for free under a Creative Commons license that allows adaption and redistribution, these online portals typically have irregularities in their markup content, or they are available exclusively in a PDF format. Web-wide pre-training corpuses such as Common Crawl might already include high-quality, open-source online textbook content, but they are generally not available in an accessible format. 

In such a scenario, online textbook websites, for example, \citet{libretexts} and \citet{openstax} are appropriate for the automatic creation of datasets for supervised learning and evaluation. OpenStax is a leading platform for open source college-level textbooks in multiple languages (English, Spanish, and Polish) in fields spanning across physical sciences, engineering, business, and life sciences. Its webpage content is more adherent to the latest online textbook standards when compared to other open source online textbooks. While there have been earlier efforts in using OpenStax as a dataset \cite{llama2-openstax}, they have focused more on the textbook content and less on end-of-chapter exercises. Web scraping the solved problem-solution pairs is also valuable for model finetuning and prompt engineering while designing language model-based student self-help tools.

In this paper, we describe the creation and application of such a dataset based on end-of-chapter solved exercises scraped from 43 college-level OpenStax textbooks in physics sciences, life sciences, mathematics, business, and social sciences, spanning 3 languages- English, Spanish, and Polish. We finetune several open source LLMs on this dataset using QLoRa \cite{dettmers2023qlora}, and evaluate its performance on the AI2RC ``challenge dev'' dataset. We also describe in detail our methodology used for scraping the web content, and discuss the challenges faced therein. We are releasing the code and dataset corresponding to this paper on Github and Huggingface respectively.\footnote{links not disclosed for anonymity during paper review} 

\section{Related Work}
Several benchmarks and evaluation methodologies have been introduced to assess the performance and capabilities of LLMs in various educational contexts. This section reviews earlier work relevant to our paper. The M3Exam benchmark is a notable example, designed to evaluate LLMs across multiple languages and modalities, including both text and images. This benchmark focuses on educational contexts and provides a comprehensive assessment of multilingual and multimodal capabilities, highlighting the importance of diverse and rich contextual information in evaluating model performance.

Similarly, the E-EVAL benchmark addresses the evaluation of LLMs within the Chinese K-12 education system. It includes a diverse set of questions across various subjects and demonstrates the process of creating and validating educational benchmarks. This study underlines the importance of subject-specific evaluation and the potential to adapt these methodologies to different educational levels and languages [34†source].

Multilingual and Multimodal Learning
The ability of LLMs to operate across multiple languages and modalities is a critical area of research. The survey "Multilingual and Multimodal Learning for Language Models" provides an extensive review of the state of the art in this field. It discusses various datasets and evaluation methods used to assess the performance of models in diverse linguistic and contextual settings. This survey emphasizes the need for benchmarks that can evaluate models in multilingual and multimodal scenarios, a need that OpenStaxQA aims to address [43†source].

Comprehensive Surveys and Benchmarks
Comprehensive surveys such as "Large Language Models for Education: A Survey and Outlook" and "Large Language Models in Education: Vision and Opportunities" offer an in-depth look into the current applications of LLMs in education. These surveys cover a wide range of topics, including benchmarks, question generation, automatic grading, and student support systems. They discuss the strengths and limitations of LLMs in generating pedagogical content and their potential impact on educational practices, providing valuable insights into the broader impacts relevant to datasets like OpenStaxQA [42†source] [34†source].

Scientific and Multilevel Evaluations
The SciEval benchmark is another important contribution, focusing on evaluating LLMs in scientific domains such as biology, chemistry, and physics. This benchmark uses a combination of static, dynamic, and experimental data to assess models' knowledge application and research abilities. The methodologies employed in SciEval can inform the development of evaluation strategies for educational content in OpenStaxQA, particularly in scientific disciplines [44†source].

Additionally, the survey "Evaluating Large Language Models: A Comprehensive Survey" provides a systematic review of various evaluation methodologies for LLMs, focusing on aspects such as knowledge, reasoning, and alignment with human values. It highlights the importance of comprehensive benchmarks for evaluating LLMs across different domains, reinforcing the need for robust evaluation frameworks like OpenStaxQA [42†source].

\section{Dataset Preparation}
In this section, we describe the web scraping methodology used in this paper for extracting the problem-solution pairs. A key differentiator for OpenStax from other major open-source online textbooks is that for the 43 textbooks we scraped in 3 languages in varied disciplines, the HTML content was consistent and well-organized into specific tags such as \verb|<os-problem-container>| (for problems) and \verb|<os-solution-container>| (for solutions). After removing non-essential HTML tags such as \verb|<div>| and \verb|<span>|, we are left with the raw text along with some MathML content. For scraping OpenStax textbooks, we used Beautiful Soup \cite{bsoup}. While we explored Scrapy \cite{scrapy}, a more advanced web scraper, we found Beautiful Soup to be sufficient for our purpose. Packages such as Scrapy are more suitable for large-scale websites with complex structures and provide features like parallel processing and request throttling, which were not necessary for scraping content from OpenStax textbooks. 

The other significant task was to convert complex MathML content into simpler formats such as Latex. This can also help us reduce the number of tokens per problem and speed up training and inference. While LLMs such as GPT-4 and ChatGPT \cite{openai2024gpt4} can  perform this task, for cost reasons we chose traditional approaches based on libraries for converting MathML to LaTeX. We evaluated XSL, Javascript, and Haskell based libraries for this purpose, and the Haskell library \verb|texmath| \cite{texmath} was found to be the most reliable in accurately converting MathML content into LaTeX content. There are cases where we might have other HTML content such as lists (\verb|<ol>|, \verb|<ul>|, etc.), and for them we use the \verb|html2text| library.

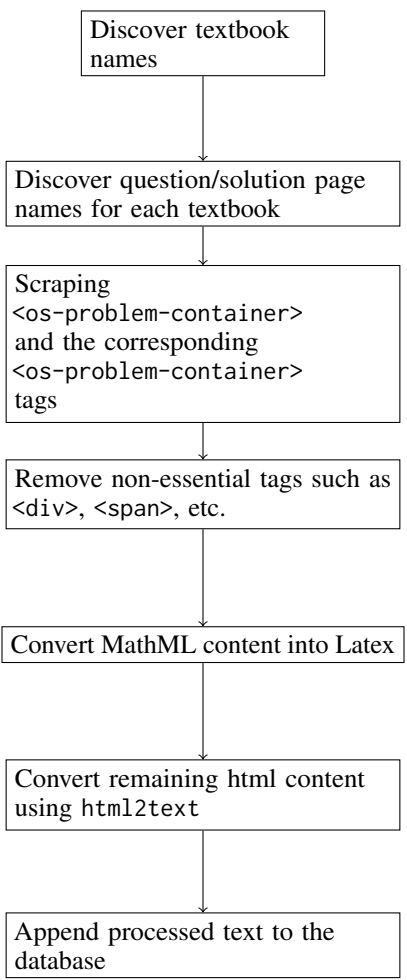
\begin{figure}[!htb]
    \centering
\begin{tikzpicture}[node distance=2cm,every node/.style={rectangle,draw}]
    % Nodes
    \node (start) [text width=3cm]{Discover textbook names};
    \node (input) [below of=start, text width=5cm] {Discover question/solution page names for each textbook};
    \node (process1) [below of=input, text width=5cm] {Scraping \verb|<os-problem-container>| and the corresponding \verb|<os-problem-container>| tags};
    \node (process2) [below of=process1, text width=5cm] {Remove non-essential tags such as \verb|<div>|, \verb|<span>|, etc.};
    \node (process3) [below of=process2] {Convert MathML content into Latex};
    \node (process4) [below of=process3, text width=5cm] {Convert remaining html content using \verb|html2text|};% Arrows
    \node (decision) [below of=process4, text width=5cm] {Append processed text to the database};% Arrows
    \draw[->] (start) -- (input);
    \draw[->] (input) -- (process1);
    \draw[->] (process1) -- (process2);
    \draw[->] (process2) -- (process3);
    \draw[->] (process3) -- (process4);
    \draw[->] (process4) -- (decision);
\end{tikzpicture}
    \caption{Scraping workflow followed for scraping Openstax question-answer pairs}
    \label{fig:scrape-flowchart}
\end{figure}

After deduplication, we finally have 18332 problem-solution pairs combined in English, Spanish, and Polish. A schematic of the dataset preparation process is described in Fig.~\ref{fig:scrape-flowchart}. Deduplication was necessary because certain textbooks have overlapping exercises-- for example, \emph{Chemistry 2e} and \emph{Chemistry 2e: atoms first}.

% \begin{pspicture}
% \sffamily
% \psset{framearc=0.3,framesep=5pt}
% \begin{psmatrix}[colsep=-0.75, rowsep=0.25]
% \psDefBoxNodes{SA}{\psframebox{\eqparbox{FC}{Sample \\Acquisition}}}\\
%    & \psDefBoxNodes{SC}{\psframebox{\eqparbox{FC}{Sample \\Checking}}} \\
%     & & \psDefBoxNodes{DSC}{\psframebox{\eqparbox{FC}{Dimensionality\\ Sample Check}}} \\
%     & & &\psDefBoxNodes{SST}{\psframebox{ \eqparbox{FC}{Sample Synthesis \\Techniques}}} \\
% \end{psmatrix}
% \psset{linecolor=Silver, doubleline, doublesep=0.35, doublecolor=Silver, arrows=->, arrowlength=0.8, arrowsize=0.6, arrowinset=0}
% \foreach \s/\t in{SA/SC, SC/DSC, DSC/SST}{\ncangle[offsetA =0.8, angleA=-90, angleB=180]{\s:bl}{\t:Cl}}
% \end{pspicture}

\section{Dataset Composition}
\begin{figure}[!htb]
    \centering
    \includegraphics[width=8cm]{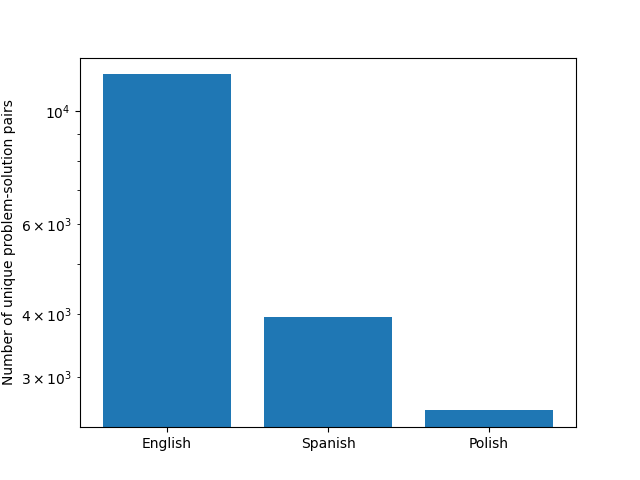}
    \caption{Language distribution in the OpenStaxQA dataset}
    \label{fig:lang-dist}
\end{figure}

As mentioned earlier, the dataset spans 3 languages- English, Spanish, and Polish. Fig.~\ref{fig:lang-dist} describes the distribution across languages. The lack of data in Polish and Spanish is because there is a lack of textbooks on OpenStax in these languages.  

\begin{figure}[!htb]
    \centering
    \includegraphics[width=8cm]{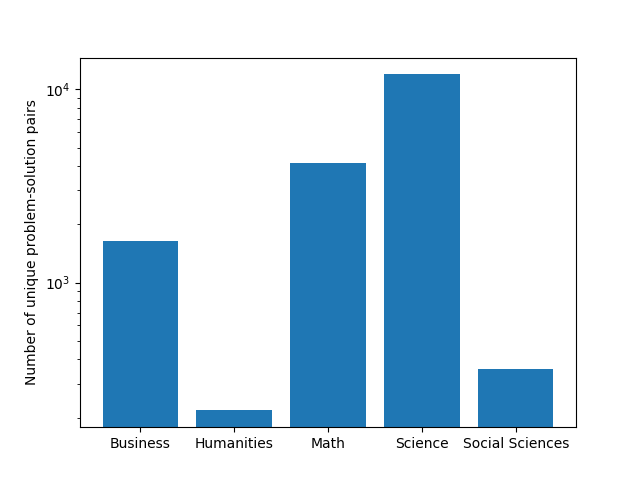}
    \caption{Distribution of fields of study in the OpenStaxQA dataset}
    \label{fig:field-dist}
\end{figure}

We also tried to represent a wide spectrum of fields of study. Fig.~\ref{fig:field-dist} describes the distribution of the fields represented in the OpenStaxQA dataset. We see that math and science fields have a higher representation compared to business, humanities, and social sciences. This is mainly because math and science questions are highly quantitative and objective, where it is easier to define a standard correct response. Whereas in humanities and social sciences, test questions are generally more subjective and do not have answers given in textbook answer keys. 

\section{Experimental Validation}
\subsection{Methodology}
Evaluation of model outputs, especially for highly technical topics covered in college courses, is complex and resource-intensive. Therefore, we decided to use GPT-4 as an oracle to rate model responses. GPT-4, given its easy accessibility and state-of-the-art text generation abilities, is a reliable proxy for human evaluations \cite{dettmers2023qlora}. The corresponding GPT-4 prompt is listed in Appendix~\ref{sec:appendix}.

\begin{table}
\centering
\begin{tabular}{p{5cm}p{2cm}}
\hline
\textbf{Rating} & \textbf{Score}\\
\hline
FULLY ACCURATE & 5 \\
MOSTLY ACCURATE & 4 \\
PARTIALLY ACCURATE & 3\\
MOSTLY INACCURATE & 2\\
FULLY INACCURATE & 1\\
\hline
\end{tabular}
\caption{\label{table:gpt4-rating}Field-wise average rating and standard deviations for the compared models}
\end{table}

\subsection{Finetuning on OpenStax dataset}
\begin{figure}[!htb]
    \centering
    \includegraphics[width=10cm]{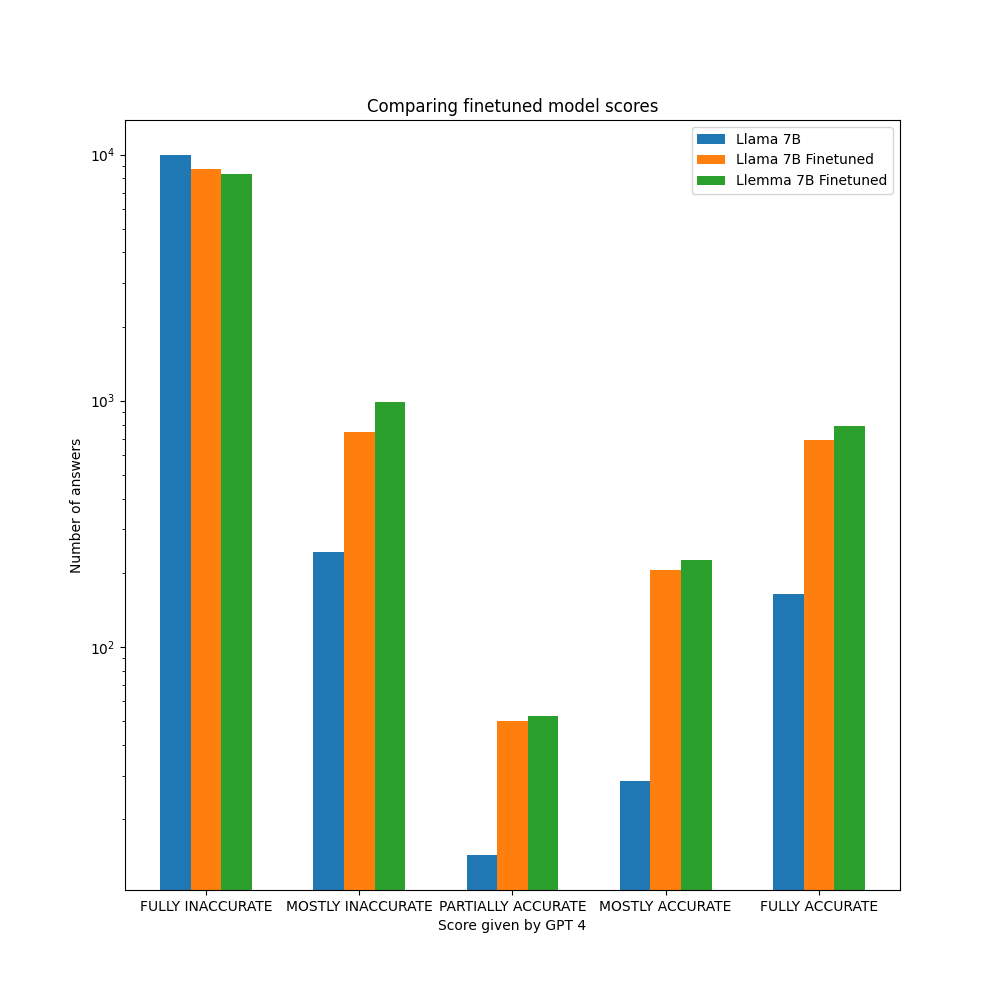}
    \caption{Finetuning results on the OpenStaxQA dataset. Comparing \emph{Llama 7B untrained}, \emph{Llama 7B finetuned}, and \emph{Llemma 7B finetuned}. Bar heights normalized for unequal sample sizes.}
    \label{fig:ft-results}
\end{figure}

% 2-sided pair t test Llama finetuned vs not finetuned TtestResult(statistic=-20.674389259534596, pvalue=7.3821465334683505e-90, df=3627.1469420289277)
% 2-sided pair t test TtestResult(statistic=-4.281653932970111, pvalue=1.8634796148322143e-05, df=20609.72120214077)
% Llama untrained- mean and standard deviations 1.0973936899862826 0.544438206075338
% Llama finetuned- mean and standard deviations 1.4066314271984623 1.0806294298169814
% Llemma finetuned- mean and standard deviations 1.472743173646876 1.1394424382483854
We create training and test datasets using a $70:30$ split, and train a QLoRA adapter for 3 epochs with 32 LLamaDecoder layers and a dropout of 0.1. We train 2 models- \verb|Llama2-7b-hf|, the 7-billion parameter Llama2 model \cite{llama2}, and \verb|Llemma-7b|, a 7-billion parameter LLM\cite{llemma} aimed at math-specific use cases, on a V100 GPU with a 32 GB GPU memory. The inference and training batch sizes were 4. The OpenStaxQA dataset has several STEM subjects which are mathematical in nature, hence we expect to observe a better performance with Llemma, as seen in the results in Table~\ref{table:ft-results-meanstdev} and Fig.~\ref{fig:ft-results}. For evaluation, we used a temperature of 1.0 and a beam size of 2. For the finetuned models, we generated 2 responses per query in the test split, in order to sample more data and reduce the uncertainty on reported metrics. We calculated the mean solution-problem token length ratio $T$, and for each batch, set a limit of $\text{floor}(T \times S)$ new tokens. Here $S$ is the length of the batch in terms of number of max tokens, and $\text{floor}(x)$ denotes the greatest integer less than equal to $x$. An interesting pattern we observe in Fig.~\ref{fig:ft-results} is that the scores provided by GPT-4 tend to be on the extremes (``fully accurate'' and ``fully inaccurate''), and there are much fewer cases that are predicted to be ``partially accurate.'' 
\begin{table}
\centering
\begin{tabular}{p{2cm}p{1.5cm}p{3cm}}
\hline
\textbf{Model} & \textbf{Mean rating (1-5)} & \textbf{paired t-test with the previous model}\\
\hline
Llama-7B & 1.10 & -\\
\addlinespace
Llama-7B finetuned & 1.41 & statistic=-20.7, p-value=$7.38 \times 10^{-90}$\\
\addlinespace
Llemma-7B finetuned & 1.47 & statistic=-4.28, p-value=$1.86 \times 10^{-5}$\\
\hline
\end{tabular}
\caption{\label{table:ft-results-meanstdev}Field-wise average rating and standard deviations for the compared models}
\end{table}

\subsection{Zero-shot evaluation on the AI2RC dataset}
\begin{figure}[!htb]
    \centering
    \includegraphics[width=10cm]{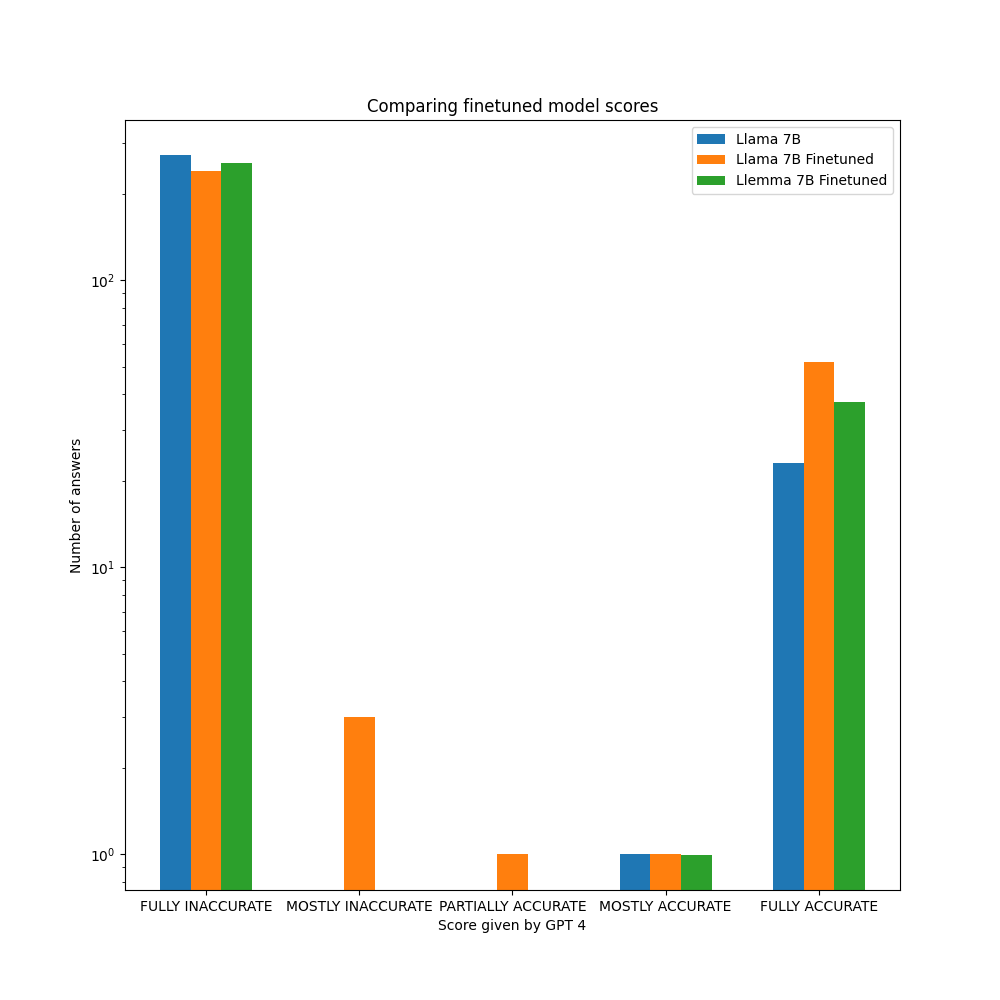}
    \caption{Finetuning results on the AI2RC dev dataset. Comparing \emph{Llama 7B untrained}, \emph{Llama 7B finetuned}, and \emph{Llemma 7B finetuned}. Bar heights normalized for unequal sample sizes.}
    \label{fig:ai2rc-ft-results}
\end{figure}

For OpenStaxQA to be a useful dataset, a model trained on it should also perform better on other education-related benchmarks. In order to test this hypothesis, we test the performance of finetuned models on the AI2RC ``challenge dev'' dataset \cite{Clark2018ThinkYH}. Table~\ref{table:ai2rc-ft-results-meanstdev} and Fig.~\ref{fig:ai2rc-ft-results} show that the finetuned models perform better than the pristine Llama 2 7B model. Moreover, we can see here that the mean rating on this dataset is better than OpenStaxQA, although the models were trained on OpenStaxQA, whereas the AI2RC evaluation is zero-shot. 

This is likely because the AI2RC dataset is purely in English and the questions are grade school level, whereas OpenStaxQA questions are college level. There could also be issues with data quality in OpenStaxQA that led to worse performance. For example, despite pre-processing, there might be residual problem-solution pairs that depend on image data or have hyperlinks to another section within the textbook. Another pattern we observe in Fig.~\ref{fig:ai2rc-ft-results}, similar to Fig.~\ref{fig:ft-results}, is that GPT-4 predicts the majority of the cases to be ``fully inaccurate'' or ``fully accurate''. Possible reasons could include the bias in GPT-4's pre-training data or Llama's pre-training data, or GPT-4's limited ability to do accurate multi-class classification. These factors could lead to a lower recall for classes such as ``mostly inaccurate,'' ``mostly inaccurate,'' and ``partially accurate.''

% 2-sided pair t test Llama finetuned vs not finetuned TtestResult(statistic=-3.7501994685340456, pvalue=0.00019603016836146795, df=533.2751988237093)
% Llama untrained- mean and standard deviations 1.3198653198653199 1.0802992393349753
% Llama finetuned- mean and standard deviations 1.7272727272727273 1.5252190778324746
\begin{table}
\centering
\begin{tabular}{p{2cm}p{1.5cm}p{3cm}}
\hline
\textbf{Model} & \textbf{Mean rating (1-5)} & \textbf{paired t-test with the previous model}\\
\hline
Llama-7B & 1.32 & -\\
\addlinespace
Llama-7B finetuned & 1.73 & statistic=-3.75, p-value=$1.96 \times 10^{-4}$\\
\addlinespace
Llemma-7B finetuned & 1.53 & statistic=1.69, p-value=0.0920\\
\hline
\end{tabular}
\caption{\label{table:ai2rc-ft-results-meanstdev}Field-wise average rating and standard deviations for the compared models, when evaluated zero-shot on the AI2RC ``challenge dev'' set. This time finetuned Llemma did worse than finetuned Llama, maybe because AI2RC questions do not have significant LaTeX content.}
\end{table}

\section{Conclusion and Outlook}
In this paper, we described OpenStaxQA, a question-answer dataset based on 43 college textbooks in 3 languages and 5 disciplines, namely, humanities, social sciences, science, math, and business. We describe the process of scraping the dataset using standard web-parser libraries and evaluate LLMs finetuned on the dataset. 

Textbooks are trustworthy and high-quality sources of concepts in higher education, and having them available for training and evaluating LLMs can lead to accelerated progress in educational applications. Traditionally, copyright issues and non-standard web interfaces have been a major impediment in this regard. However, as more authors make their published textbook content in standardized web interfaces under permissive licenses such as Creative Commons, there is a need to also standardize the process of converting the complex HTML/XML/MathML content of textbook webpages into datasets for training NLP-based tools for educational applications. Different open-source textbook platforms have different formats, and in many cases, redundant and erroneous HTML content, insufficiently tagged textbook content, and non-standard organization of chapters makes it hard for a software developer or researcher to utilize a textbook for training and evaluating models. We publicly release our web scraping code, so that researchers scraping educational content from other textbooks can build on top of it, with the hope that eventually we will have robust tools that allow scraping textbook content at scale. This works the other way too -- existing computer vision and NLP tools can be used to parse textbook content that is difficult to parse using traditional rule-based parsers. This also extends to open source textbooks that are only available in PDF format. 

Another opportunity is for recent STEM textbooks whose LaTeX version is made available under permissive licenses by the authors. Diversity in terms of content and languages is a major issue in NLP-based educational applications, and we need to build datasets for college and graduate level textbooks in other countries and languages. We plan to take up these directions as an extension to this work, by making use of other open source textbook platforms such as OER Commons and Open Textbook Library, which are available in diverse subjects and languages.

% we will scrape more languages from the open textbook thing in UMich

\section*{Limitations}
Large language models reflect the biases in their training data \cite{Schramowski2021LargePL}, and this bias can also reflect in models finetuned on datasets such as OpenStaxQA. In a review of the ethical limitations of educational tools based on LLMs, \citet{Yan2023_llm_education_impact} describe privacy, transparency, beneficence, and equality issues. For example, access to an educational tool finetuned on datasets such as OpenStaxQA might be appropriate for a student when they review course concepts, but not when they have a take-home test. Similarly, students with stronger financial backgrounds and well-versed in high-resource languages such as English tend to benefit more from such tools than students who do not have these privileges.

% Table~\ref{citation-guide} shows the syntax supported by the style files.
% We encourage you to use the natbib styles.
% You can use the command \verb|\citet| (cite in text) to get ``author (year)'' citations, like this citation to a paper by \citet{Gusfield:97}.
% You can use the command \verb|\citep| (cite in parentheses) to get ``(author, year)'' citations \citep{Gusfield:97}.
% You can use the command \verb|\citealp| (alternative cite without parentheses) to get ``author, year'' citations, which is useful for using citations within parentheses (e.g. \citealp{Gusfield:97}).

% A possessive citation can be made with the command \verb|\citeposs|.
% This is not a standard natbib command, so it is generally not compatible with other style files.

% \section*{Acknowledgements}
% We would like to acknowledge OpenStax and all the corresponding textbook authors for their online textbook repository available under a CC-BY-4.0 Creative Commons Attribution License \cite{creative-commons-license}. We thank Dr. Colin Clement for their suggestions regarding dataset evaluations.

\bibliographystyle{plainnat}
\bibliography{custom}

\begin{thebibliography}{14}
\providecommand{\natexlab}[1]{#1}
\providecommand{\url}[1]{\texttt{#1}}
\expandafter\ifx\csname urlstyle\endcsname\relax
  \providecommand{\doi}[1]{doi: #1}\else
  \providecommand{\doi}{doi: \begingroup \urlstyle{rm}\Url}\fi

\bibitem[Azerbayev et~al.(2023)Azerbayev, Schoelkopf, Paster, Santos, McAleer, Jiang, Deng, Biderman, and Welleck]{llemma}
Zhangir Azerbayev, Hailey Schoelkopf, Keiran Paster, Marco~Dos Santos, Stephen McAleer, Albert~Q. Jiang, Jia Deng, Stella Biderman, and Sean Welleck.
\newblock Llemma: An open language model for mathematics, 2023.

\bibitem[Beeching et~al.(2023)Beeching, Belkada, Rasul, Tunstall, von Werra, Rajani, and Lambert]{beeching2023stackllama}
Edward Beeching, Younes Belkada, Kashif Rasul, Lewis Tunstall, Leandro von Werra, Nazneen Rajani, and Nathan Lambert.
\newblock Stackllama: An rl fine-tuned llama model for stack exchange question and answering, 2023.
\newblock URL \url{https://huggingface.co/blog/stackllama}.

\bibitem[Clark et~al.(2018)Clark, Cowhey, Etzioni, Khot, Sabharwal, Schoenick, and Tafjord]{Clark2018ThinkYH}
Peter Clark, Isaac Cowhey, Oren Etzioni, Tushar Khot, Ashish Sabharwal, Carissa Schoenick, and Oyvind Tafjord.
\newblock Think you have solved question answering? try arc, the ai2 reasoning challenge.
\newblock \emph{ArXiv}, abs/1803.05457, 2018.
\newblock URL \url{https://api.semanticscholar.org/CorpusID:3922816}.

\bibitem[Dettmers et~al.(2023)Dettmers, Pagnoni, Holtzman, and Zettlemoyer]{dettmers2023qlora}
Tim Dettmers, Artidoro Pagnoni, Ari Holtzman, and Luke Zettlemoyer.
\newblock Qlora: Efficient finetuning of quantized llms, 2023.

\bibitem[{Fahrial Fiansyah}(2024)]{llama2-openstax}
{Fahrial Fiansyah}.
\newblock {Llama 2 Openstax Huggingface Model}.
\newblock {https://huggingface.co/fahrialfiansyah/llama-2-13b-openstax-v2}, February 2024.
\newblock {Accessed: March 5, 2024}.

\bibitem[Kouzis-Loukas(2016)]{scrapy}
Dimitrios Kouzis-Loukas.
\newblock \emph{Learning Scrapy}.
\newblock Packt Publishing Ltd, 2016.

\bibitem[{Libre Texts}(2024)]{libretexts}
{Libre Texts}.
\newblock {Online Textbook Library}.
\newblock {https://libretexts.org}, February 2024.
\newblock {Accessed: March 6, 2024}.

\bibitem[McFarlane(2024)]{texmath}
Joe McFarlane.
\newblock Texmath library, 2024.
\newblock URL \url{https://github.com/jgm/texmath}.

\bibitem[OpenAI et~al.(2024)OpenAI, Achiam, Adler, Agarwal, Ahmad, et~al.]{openai2024gpt4}
OpenAI, Josh Achiam, Steven Adler, Sandhini Agarwal, Lama Ahmad, et~al.
\newblock Gpt-4 technical report, 2024.

\bibitem[{OpenStax}(2024)]{openstax}
{OpenStax}.
\newblock {Online Textbook Library}.
\newblock {https://openstax.org}, February 2024.
\newblock {Accessed: March 5, 2024}.

\bibitem[Richardson(2007)]{bsoup}
Leonard Richardson.
\newblock {Beautiful Soup Documentation}, 2007.
\newblock URL \url{https://www.crummy.com/software/BeautifulSoup/bs4/doc/}.

\bibitem[Schramowski et~al.(2021)Schramowski, Turan, Andersen, Rothkopf, and Kersting]{Schramowski2021LargePL}
Patrick Schramowski, Cigdem Turan, Nico Andersen, Constantin~A. Rothkopf, and Kristian Kersting.
\newblock Large pre-trained language models contain human-like biases of what is right and wrong to do.
\newblock \emph{Nature Machine Intelligence}, 4:\penalty0 258 -- 268, 2021.
\newblock URL \url{https://api.semanticscholar.org/CorpusID:246824056}.

\bibitem[{Touvron} et~al.(2023){Touvron}, {Martin}, {Stone}, {Albert}, {Almahairi}, {Babaei}, {Bashlykov}, {Batra}, {Bhargava}, {Bhosale}, {Bikel}, {Blecher}, {Canton Ferrer}, {Chen}, {Cucurull}, {Esiobu}, {Fernandes}, {Fu}, {Fu}, {Fuller}, {Gao}, {Goswami}, {Goyal}, {Hartshorn}, {Hosseini}, {Hou}, {Inan}, {Kardas}, {Kerkez}, {Khabsa}, {Kloumann}, {Korenev}, {Singh Koura}, {Lachaux}, {Lavril}, {Lee}, {Liskovich}, {Lu}, {Mao}, {Martinet}, {Mihaylov}, {Mishra}, {Molybog}, {Nie}, {Poulton}, {Reizenstein}, {Rungta}, {Saladi}, {Schelten}, {Silva}, {Smith}, {Subramanian}, {Tan}, {Tang}, {Taylor}, {Williams}, {Kuan}, {Xu}, {Yan}, {Zarov}, {Zhang}, {Fan}, {Kambadur}, {Narang}, {Rodriguez}, {Stojnic}, {Edunov}, and {Scialom}]{llama2}
Hugo {Touvron}, Louis {Martin}, Kevin {Stone}, Peter {Albert}, Amjad {Almahairi}, Yasmine {Babaei}, Nikolay {Bashlykov}, Soumya {Batra}, Prajjwal {Bhargava}, Shruti {Bhosale}, Dan {Bikel}, Lukas {Blecher}, Cristian {Canton Ferrer}, Moya {Chen}, Guillem {Cucurull}, David {Esiobu}, Jude {Fernandes}, Jeremy {Fu}, Wenyin {Fu}, Brian {Fuller}, Cynthia {Gao}, Vedanuj {Goswami}, Naman {Goyal}, Anthony {Hartshorn}, Saghar {Hosseini}, Rui {Hou}, Hakan {Inan}, Marcin {Kardas}, Viktor {Kerkez}, Madian {Khabsa}, Isabel {Kloumann}, Artem {Korenev}, Punit {Singh Koura}, Marie-Anne {Lachaux}, Thibaut {Lavril}, Jenya {Lee}, Diana {Liskovich}, Yinghai {Lu}, Yuning {Mao}, Xavier {Martinet}, Todor {Mihaylov}, Pushkar {Mishra}, Igor {Molybog}, Yixin {Nie}, Andrew {Poulton}, Jeremy {Reizenstein}, Rashi {Rungta}, Kalyan {Saladi}, Alan {Schelten}, Ruan {Silva}, Eric~Michael {Smith}, Ranjan {Subramanian}, Xiaoqing~Ellen {Tan}, Binh {Tang}, Ross {Taylor}, Adina {Williams}, Jian~Xiang {Kuan}, Puxin {Xu}, Zheng {Yan}, Iliyan {Zarov},
  Yuchen {Zhang}, Angela {Fan}, Melanie {Kambadur}, Sharan {Narang}, Aurelien {Rodriguez}, Robert {Stojnic}, Sergey {Edunov}, and Thomas {Scialom}.
\newblock {Llama 2: Open Foundation and Fine-Tuned Chat Models}.
\newblock \emph{arXiv e-prints}, art. arXiv:2307.09288, July 2023.
\newblock \doi{10.48550/arXiv.2307.09288}.

\bibitem[Yan et~al.(2023)Yan, Sha, Zhao, Li, Martinez‐Maldonado, Chen, Li, Jin, and Gašević]{Yan2023_llm_education_impact}
Lixiang Yan, Lele Sha, Linxuan Zhao, Yuheng Li, Roberto Martinez‐Maldonado, Guanliang Chen, Xinyu Li, Yueqiao Jin, and Dragan Gašević.
\newblock Practical and ethical challenges of large language models in education: A systematic scoping review.
\newblock \emph{British Journal of Educational Technology}, 55\penalty0 (1):\penalty0 90–112, August 2023.
\newblock ISSN 1467-8535.
\newblock \doi{10.1111/bjet.13370}.
\newblock URL \url{http://dx.doi.org/10.1111/bjet.13370}.

\end{thebibliography}

\appendix

\section{GPT-4 Prompt}
\label{sec:appendix}
We used the following prompt for rating the model responses into 5 distinct possibilities, as listed in Table~\ref{table:gpt4-rating}:
\begin{quote}
You are evaluating solutions to textbook problems given by a language model. 
Given a 'PROBLEM', 'CORRECT SOLUTION', and 'LANGUAGE MODEL RESPONSE', you rate the 'LANGUAGE MODEL RESPONSE' and detect whether it is a satisfactory response to the 'PROBLEM'. 
Your answer should be a single rating from 5 options: ['FULLY ACCURATE', 'MOSTLY ACCURATE', 'PARTIALLY ACCURATE', 'MOSTLY INACCURATE', 'FULLY INACCURATE'] . 
It does not matter whether 'LANGUAGE MODEL RESPONSE' is verbose or terse, only its accuracy matters.
\end{quote}
\end{document}